\documentclass[conference]{IEEEtran}
\IEEEoverridecommandlockouts
\usepackage{cite}
\usepackage{amsmath,amssymb,amsfonts}
\usepackage{algorithmic}
\usepackage{graphicx}
\usepackage[whole]{bxcjkjatype}
\usepackage{url}
\usepackage{textcomp}
\usepackage{xcolor}
\def\BibTeX{{\rm B\kern-.05em{\sc i\kern-.025em b}\kern-.08em
    T\kern-.1667em\lower.7ex\hbox{E}\kern-.125emX}}

\usepackage{tcolorbox}
\tcbuselibrary{raster,skins,breakable}
\tcbuselibrary{xparse} 
\DeclareTColorBox{brekableitembox}{ o m O{.5} O{} }%
  {empty, left=2mm, right=2mm, top=-2mm, bottom=0.2mm, attach boxed title to top left={xshift=1.2em},
  boxed title style={empty,left=-2mm,right=-2mm}, colframe=black, coltitle=black, coltext=black, breakable,  
  underlay unbroken={\draw[black,line width=#3pt]
    (title.east) -- (title.east-|frame.east) -- (frame.south east) -- (frame.south west) -- (title.west-|frame.west) -- (title.west); },
  underlay first={\draw[black,line width=#3pt](title.east) -- (title.east-|frame.east) -- (frame.south east) ;
    \draw[black,line width=#3pt] (frame.south west) -- (title.west-|frame.west) -- (title.west); },
  underlay middle={\draw[black,line width=#3pt](frame.north east) -- (frame.south east) ;
    \draw[black,line width=#3pt](frame.south west) -- (frame.north west) ;},
  underlay last={\draw[black,line width=#3pt](frame.north east) -- (frame.south east) -- (frame.south west) -- (frame.north west) ;}, IfValueTF={#1}{title=~~#2~~〈#1〉}{title=~~#2~~},#4}
\begin{document}

\title{Enhancing Financial Domain Adaptation of Language Models via Model Augmentation
}

\author{\IEEEauthorblockN{Kota Tanabe}
\IEEEauthorblockA{\textit{Faculty of Information Science and Technology} \\
\textit{Hokkaido University}\\
0009-0005-5979-2043}
\and
\IEEEauthorblockN{Masanori Hirano}
\IEEEauthorblockA{
\textit{Preferred Networks, Inc.}\\
Tokyo, Japan \\
research@mhirano.jp}
\and
\IEEEauthorblockN{Kazuki Matoya}
\IEEEauthorblockA{
\textit{Preferred Networks, Inc.}\\
Tokyo, Japan \\
kmatoya@preferred.jp}
\and
\IEEEauthorblockN{Kentaro Imajo}
\IEEEauthorblockA{
\textit{Preferred Networks, Inc.}\\
Tokyo, Japan \\
imos@preferred.jp}
\and
\IEEEauthorblockN{Hiroki Sakaji}
\IEEEauthorblockA{\textit{Faculty of Information Science and Technology} \\
\textit{Hokkaido University}\\
0000-0001-5030-625X}
\and
\IEEEauthorblockN{Itsuki Noda}
\IEEEauthorblockA{\textit{Faculty of Information Science and Technology} \\
\textit{Hokkaido University}\\
0000-0003-1987-5336}
}

\maketitle

\begin{abstract}
The domain adaptation of language models, including large language models (LLMs), has become increasingly important as the use of such models continues to expand. This study demonstrates the effectiveness of Composition to Augment Language Models (CALM) in adapting to the financial domain. 
CALM is a model to extend the capabilities of existing models by introducing cross-attention between two LLMs with different functions. In our experiments, we developed a CALM to enhance the financial performance of an LLM with strong response capabilities by leveraging a financial-specialized LLM. Notably, the CALM was trained using a financial dataset different from the one used to train the financial-specialized LLM, confirming CALM's ability to adapt to various datasets. The models were evaluated through quantitative Japanese financial benchmarks and qualitative response comparisons, demonstrating that CALM enables superior responses with higher scores than the original models and baselines. Additionally, comparative experiments on connection points revealed that connecting the middle layers of the models is most effective in facilitating adaptation to the financial domain. These findings confirm that CALM is a practical approach for adapting LLMs to the financial domain.
\end{abstract}

\begin{IEEEkeywords}
Financial Natural Language Processing, Large Language Model, Domain Adaptation
\end{IEEEkeywords}

\section{Introduction}

We are witnessing a revolutionary shift in natural language processing (NLP) due to advancements in generative AI technologies, such as ChatGPT. These generative AI models, including large language models (LLMs), are being actively developed and utilized by researchers, companies, and individuals. As these models become more widespread, domain adaptation has emerged as a critical issue for their practical application across diverse fields and contexts.
One prominent domain where LLMs demonstrate significant potential is finance. In the financial sector, LLMs offer promising capabilities for various tasks, including information extraction, summarization, and classification of financial documents. This potential has led to a growing demand for specialized financial LLMs. Notable examples include BloombergGPT~\cite{wu2023bloomberggpt}, FinGPT~\cite{yang2023fingpt}, and nekomata-14b-pfn-qfin~\cite{hirano2024construction}, all of which are trained on extensive financial datasets to enhance their domain-specific performance.

However, LLMs adapted to the financial domain often excel in financial knowledge and understanding but may lack strong reasoning or language generation capabilities. Conversely, models with robust general language understanding and generation abilities may not be well-suited for financial tasks due to their lack of domain-specific adaptation. Developing a model that successfully integrates both capabilities remains a significant challenge. A typical approach to address this is to continue pre-training or fine-tuning a model that excels in language understanding and generation using financial datasets. However, this approach is often challenging to implement due to the high computational costs of training LLMs with a large number of parameters.

In this context, Composition to Augment Language Models (CALM)~\cite{bansal2024llm}, which enhances domain adaptation by connecting an anchor model with a domain-specific model, has gained significant attention. CALM enables the scaling up of existing models without altering their parameters, using only a small number of additional parameters and a limited dataset to connect different LLMs. The effectiveness of this approach has been demonstrated in various tasks such as low-resource language translation, arithmetic reasoning, and code explanation/generation. Additionally, CALM is considered advantageous because it avoids the catastrophic forgetting often associated with traditional approaches involving further pre-training. However, a notable limitation of CALM is that the dataset used for its training is typically a subset of the data used to build the augmenting model. This can pose challenges when the financial datasets used to construct the models are proprietary to an organization, making it difficult for external users to access and leverage CALM.

In this study, we aim to facilitate the adaptation of an anchor model to the financial domain by utilizing a model specifically tailored for financial knowledge and understanding. We demonstrate that models constructed using CALM outperform baseline models in financial domain adaptation, as evidenced by both quantitative evaluations using financial benchmarks and qualitative assessments through comparative analysis of model outputs. The contributions of this study are as follows: (1) We show that CALM is effective in adapting models to the financial domain. (2) We also demonstrate that CALM remains effective even when the financial dataset used for training differs from the dataset used to build the augmenting model.

\section{Related Work}
Methodologies to fit language models to specific domains have been pursued.
For instance, Howard {\it et al.} \cite{howard2018ulmfit} proposed universal language model fine-tuning, following which, some domain-specific language models were developed, such as SciBERT \cite{beltagy2019scibert}, MedBERT \cite{rasmy2021med}, FinBERT \cite{huang2023finbert,liu2021finbert}, and Japanese financial BERT \cite{Suzuki2022-sigfin28}.
Moreover, the methodologies and effects of domain-specified fine-tuning were discussed in \cite{gururangan2020don,Suzuku2023-ipm}.

In the LLM era, although several transformer-based language models have been proposed, as described in the Introduction, several unknown LLM mechanisms exist and numerous trials have been performed.

Several studies proposed LLMs that focus specifically on finance using continual pretraining.
For instance, BloombergGPT \cite{Wu2023} is a private LLM focused on finance.
In addition, publicly available models, such as FinLLAMA \cite{Fin-LLAMA}, which is a tuned version of LLaMA \cite{touvron2023llama}, FinGPT \cite{yang2023fingpt}, Instruct-FinGPT \cite{zhang2023instruct}, and LLaMA-2-Econ \cite{keles2024llama} also exist.
However, instruction-tuned LLMs for finance are scarce and, in the study of Instruct-FinGPT \cite{zhang2023instruct}, only support sentiment analysis tasks.
In this study, we utilize nekomata-14b-pfn-qfin \cite{hirano2024construction}, which focuses on the Japanese and financial domains and is built by the Japanese continual pretraining for the financial domain.

Other than continual pretraining, several methods to fit LLMs to the specific domain are proposed.
Model merging on LLMs is one example.
Ilharco {\it et al.} \cite{ilharco2022editing} demonstrated task arithmetic model merging, merging models with different task-solving skills.
Additionally, Anthropic reports some interpretable features in LLMs and the possibility of controlling the model outputs by the strength of each feature \cite{templeton2024scaling}, which could prove the effectiveness of task arithmetic.
More complex model merging, such as ties-merging \cite{yadav2024ties}, DARE \cite{zhao2024towards}, and evolutionary model merging \cite{akiba2024evolutionary}, is also proposed.
In this study, we employed CALM \cite{bansal2024llm}, which is a different method to fit LLMs to specific domains or tasks by bridging multiple LLMs.

\section{CALM Details}

CALM aims to synthesize new capabilities by combining the strengths of an anchor model $\mathrm{m_B}$ and an augmenting model $\mathrm{m_A}$. In this setup, $\mathrm{m_A}$ possesses domain-specific abilities that $\mathrm{m_B}$ lacks, while $\mathrm{m_B}$ is a model known for its high language generation and reasoning capabilities. The connection between $\mathrm{m_B}$ and $\mathrm{m_A}$ is established through newly added parameters, with the weights of both models remaining frozen. Only the additional parameters are updated during training. The dataset used for training includes a subset of the data that was originally used to endow $\mathrm{m_A}$ with its specific knowledge or skills.

Specifically, additional parameters are introduced and trained on selected layers from both $\mathrm{m_A}$ and $\mathrm{m_B}$: (1) a linear transformation $f_{proj}(.)$ that projects the $i^{th}$ layer representation $H_{Ai}$ of $\mathrm{m_A}$ into the dimensional space of the $j^{th}$ layer representation $H_{Bj}$ of $\mathrm{m_B}$, and (2) the parameters of a cross-attention layer $f_{cross}(,.,)$ that leverages the transformed representation and $H_{Bj}$. In this study, since the architectures of $\mathrm{m_B}$ and $\mathrm{m_A}$ are identical, and the dimension $D$ of their representations is the same, $f_{proj}(.)$ is not introduced. The operation of the cross-attention can be expressed as follows:


\begin{align*}
f_{\text{cross}}(H_{Ai}, H_{Bj}) &= \text{Concat.}_{k}(head_{k})W^O \\
\text{where,} \quad head_k &= \text{Attn}(Q_B, K_A, V_A) \\
\text{and,} \quad Q_B &= H_{Bj}W^Q_{k}, \\
K_A &= H_{Ai}W^{K}_{k}, \\
V_A &= H_{Ai}W^{V}_{k}
\end{align*}

Here, $N_H$ denotes the number of heads in the Multi-head Attention, with $k \in {1..N_H}$, and the weight matrices $W^O \in \mathbb{R}^{D \times D}$, $W^{K}{k}, W^{V}{k}, W^{Q}_{k} \in \mathbb{R}^{D \times \frac{D}{N_H}}$ are the parameters to be learned. The output of the cross-attention is then added as a residual connection to the layer representation of $\mathrm{m_B}$:

\begin{align*}
\mathbf{H}_{A_i \oplus B_j} = \mathbf{H}_{B_j} + f_{\text{cross}}\left(\mathbf{H}_{A_i}, \mathbf{H}_{B_j} \right)
\end{align*}
Here, $\mathbf{H}_{A_i \oplus B_j}$ represents the input to the $(j+1)^{th}$ layer of $\mathrm{m_B}$. Following this operation, the output token $y_t$ at time step $t$ is obtained from $\mathrm{m_B}$. This token $y_t$ is subsequently used as input to both $\mathrm{m_A}$ and $\mathrm{m_B}$ at the next time step. Consequently, a sequence of tokens is generated sequentially, similar to a typical autoregressive language model.

\section{Experiment}
In this study, we aim to facilitate the adaptation of the anchor model to the financial domain by leveraging the financial knowledge of the augmenting model.
\subsection{Models}
This study utilizes two models from the nekomata family, a series of Japanese language models. The anchor model is nekomata-14b-instruction\footnote{\url{https://huggingface.co/rinna/nekomata-14b-instruction}}, released by Rinna Co., Ltd.\footnote{\url{https://rinna.co.jp/}}. The augmenting model is nekomata-14b-pfn-qfin~\cite{hirano2024construction}, provided by Preferred Networks, Inc.\footnote{\url{https://www.preferred.jp/ja/}}.
Nekomata-14b-instruction is derived from nekomata-14b, which was created through continued pre-training on Japanese language data using Qwen-14B~\cite{qwen} as a base. The instruction model was obtained by applying Instruction Tuning to nekomata-14b, enhancing its ability to follow user's instructions.
On the other hand, nekomata-14b-pfn-qfin is a financial language model. It was created by further pre-training nekomata-14b with a proprietary financial dataset collected from the internet. This dataset consists of approximately 8.1 million documents and 370 million tokens, sourced from finance-specific texts such as minutes of central bank policy meetings and reports published by various financial institutions.

\subsection{Dataset}
The Japanese Financial Instruction Dataset~\cite{tanabe2024jafin} was used for training CALM. This dataset is distinct from the financial dataset employed in the domain adaptation of the augmenting model. It contains 1,490 question-and-answer samples encompassing financial knowledge. The dataset was primarily collected from frequently asked questions and explanatory materials on financial-related websites of Japanese government entities such as the Ministry of Finance, the Japan Pension Service, and the Financial Services Agency, as well as from Wikipedia pages related to finance. It includes knowledge on a wide range of topics, such as financial policy, taxation, investment, and the pension system.

\subsection{CALM settings}
In the experiment, the dataset was divided into training and validation sets with a ratio of 85\% to 15\%, and only the connection parts constructed with cross-attention were targeted for training. The number of cross-attention heads was set to 128. Both the anchor and augmenting models consist of 40 layers, and we evaluated their performance on financial tasks when the connection interval was set at every four layers, as well as the performance of each model independently. Furthermore, to examine the impact of connection points on financial task performance, we also compared cases where a single connection was made at the models' head, middle, or tail. Additionally, the performance of the anchor model was compared with the results of training using LoRA~\cite{hu2021lora}, a representative fine-tuning method for language models.

The training with LoRA and CALM was conducted for 20 epochs, and the model obtained at the epoch with the lowest validation loss was evaluated for its adaptation to the financial domain. To ensure comparability, the rank of LoRA was set to match the number of additional parameters introduced by CALM.

\subsection{Evaluation}
Each model is evaluated using a benchmark designed to measure the generation quality of LLMs in Japanese financial conversations~\cite{Hirano2024-pfmt}. The benchmark consists of 360 dialogues, each comprising two turns. The tasks cover 12 categories: writing, roleplay, knowledge, extraction, reasoning, math, coding, idea, translation, ethics, trustworthiness, and ESGs. Evaluation is conducted on a 10-point scale by GPT-4.
\section{Result and Discussion}
\subsection{Quantitative Evaluation}

The results of the evaluation experiments are presented in Table \ref{tab:results}. The scores of CALM surpassed those of both the anchor model and the augmenting model when evaluated individually, in both the First turn and Second turn. This indicates that CALM effectively enhances the performance of the anchor model in the financial domain by leveraging the financial knowledge embedded in the augmenting model. This improvement is likely due to the cross-attention mechanism facilitating effective integration between the models, enabling better adaptation to the financial domain. Through cross-attention, the anchor model selectively incorporates financial knowledge from the augmenting model, rather than directly using its representations. This selective incorporation allows the generation of high-quality responses grounded in sophisticated financial knowledge. Although the scores of LoRA exceeded those of the individual anchor and augmenting models, they were lower than those achieved by CALM. This suggests that CALM is more effective than LoRA in adapting models to the financial domain. One of the primary reasons CALM outperformed LoRA is likely because of CALM's ability to access a broader range of financial knowledge compared to LoRA.
The financial knowledge in the datasets used for training is limited, which may result in the LoRA-trained model being less capable of responding appropriately in unfamiliar contexts. In contrast, CALM updates its parameters not by embedding financial knowledge into additional parameters, but by facilitating a more seamless integration with the augmenting model, which possesses extensive financial expertise. Consequently, CALM is able to adapt to a wider range of tasks in the financial domain without relying solely on the limited knowledge of the training dataset, thereby demonstrating superior performance compared to LoRA. 
Furthermore, Table \ref{tab:results-connect-one} compares the scores when only one connection is made at the model's head, middle, or tail. The results show that the model with the connection at the middle layer achieved the highest score, followed by the model with the connection at the tail, and then the model with the connection at the head. These findings suggest that connecting the middle layers of the models is most effective for adapting the anchor model to the financial domain. This effectiveness can be attributed to the fact that the middle layers capture features obtained from earlier processing stages while being sufficiently distant from the final layers, where text generation is finalized. This allows for information to be represented at an optimal level of abstraction, facilitating the integration of features.



\begin{figure*}[tbp]
\centerline{\includegraphics[width=0.9\textwidth]{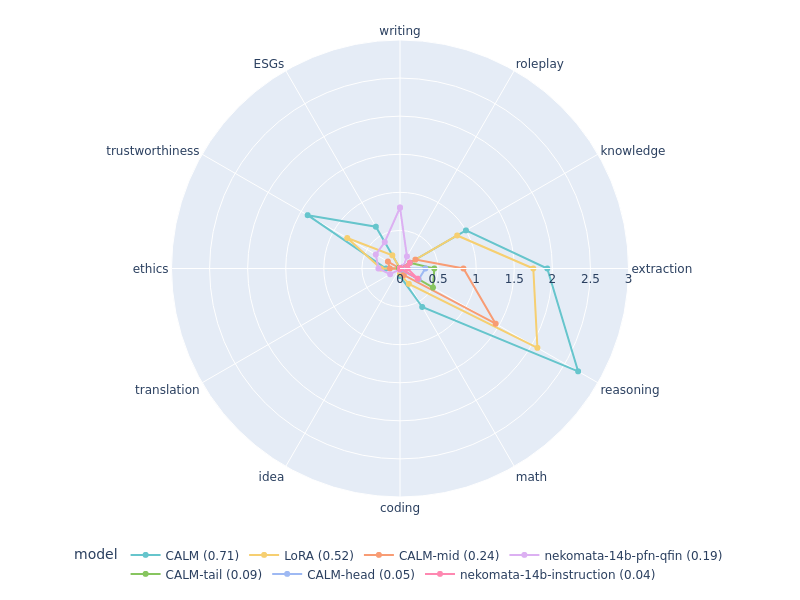}}
\caption{Radar chart comparing the scores of each model across different tasks on the Japanese financial benchmark}
\label{rader}
\end{figure*}

The performance comparison for each task across the different models is shown in Fig.~\ref{rader}. The radar chart shows that the scores for tasks such as knowledge, extraction, reasoning, math, trustworthiness, and ESGs have significantly improved compared to the anchor and augmenting model individually.
Compared to LoRA, the model slightly underperformed in the ethics task but matched or exceeded LoRA's performance in all other tasks. This suggests that the augmenting model can provide the anchor model with a broader range of financial knowledge than what is contained in the financial dataset used for training.
However, the scores were either 0 or close to 0 in half of the remaining tasks. This issue appears to stem from a characteristic of the anchor and augmenting models, where the model generates the \texttt{<endoftext>} token at the beginning of the output, preventing the generation of subsequent tokens. While the cross-attention connections effectively enhance the anchor model, CALM may also inherit such unfavorable traits from individual models. Nevertheless, despite this drawback, the superior scores of CALM in half of the tasks suggest that the financial knowledge embedded in the augmenting model has strongly facilitated the adaptation of the anchor model to the financial domain.
For the model with only one connection point, the central connection outperformed the front and back connections across all tasks, indicating that the central connection may be the most effective regardless of the task type. The consistently high performance of the central connection indicates that CALM may not require adjustments to the connection layer for different tasks. The ability to handle various financial tasks with a consistent structure suggests that this method could be highly practical for financial domain adaptation.

\begin{table*}[htbp]
\caption{A comparison of scores on the Japanese financial benchmark between CALM, nekomata-14b-pfn-qfin (augmenting model only), nekomata-14b-instruction (anchor model only), and a model where LoRA is applied to nekomata-14b-instruction using a Japanese financial instruction dataset.}
\centering
\renewcommand{\arraystretch}{1.5}
\resizebox{0.6\linewidth}{!}{%
\begin{tabular}{lcccccc}
\hline
\textbf{Model} & \textbf{Ave.} & \textbf{First turn} & \textbf{Second turn} \\
\hline
augmenting model only & 0.187500 & 0.227778 & 0.147222\\
anchor model only & 0.043056 & 0.041667& 0.044444\\
LoRA & 0.520833 & 0.488889 & 0.552778\\
CALM & \textbf{0.711111} & \textbf{0.627778} & \textbf{0.794444}\\
\hline
\end{tabular}%
}
\label{tab:results}
\end{table*}

\begin{table*}[htbp]
\caption{Comparison of scores based on the connection points when connecting only one layer between the Anchor and Augmenting models."-head," "-mid," and "-tail" refer to the use of outputs from the 0th layer, 20th layer, and 38th layer of the Anchor and Augmenting models, respectively, to construct cross-attention.}
\centering
\renewcommand{\arraystretch}{1.5}
\resizebox{0.52\linewidth}{!}{%
\begin{tabular}{lcccccc}
\hline
\textbf{Model} & \textbf{Ave.} & \textbf{First turn} & \textbf{Second turn} \\
\hline
CALM-head & 0.052778 & 0.061111 & 0.044444\\
CALM-mid & \textbf{0.244444} & \textbf{0.258333} & \textbf{0.230556}\\
CALM-tail & 0.093056 & 0.102778 & 0.083333\\
\hline
\end{tabular}%
}
\label{tab:results-connect-one}
\end{table*}


\subsection{Qualitative Evaluation}
We compare the outputs of each model in response to the benchmark inputs below. If the output contains only the \texttt{<endoftext>} token, it indicates that the model generated only \texttt{<endoftext>}, causing the generation process to stop. In this section, we present the English translations of the outputs, while the original Japanese text is provided in the Appendix.







\begin{brekableitembox}{Response Example 1}
Compare the characteristics of the following financial products, choose the one you believe is most suitable for long-term asset building, and explain your reasoning.\\
a) Stocks: High returns can be expected, but they carry a high risk of price volatility.\\
b) Bonds: Provide stable interest income, but are subject to interest rate risk.\\
c) Investment Funds: Risk diversification is achieved through diversified investments, but management costs are incurred.\\
d) Deposits: Principal is guaranteed, but returns are limited in a low-interest-rate environment.\\

(augmenting model only) \\
\texttt{<endoftext>} \\

(anchor model only) \\
a) Stocks can be expected to yield high returns, but they also carry a high risk of price volatility, which may lead to short-term losses. However, over the long term, the stock market generally tends to rise due to corporate growth and overall economic growth. Therefore, stocks can be considered the most suitable option for long-term asset building.\\

(LoRA) \\
a) Stocks can be expected to yield high returns, but they also carry a high risk of price volatility, which may lead to short-term losses. However, over the long term, the value of stocks is expected to increase due to corporate growth and overall economic growth. Therefore, stocks are considered the most suitable option for long-term asset building.\\

(CALM) \\
a) Stocks can be expected to yield high returns, but they also carry a high risk of price volatility, which may result in short-term losses. However, over the long term, stock prices have the potential to increase due to corporate growth and economic conditions, making them suitable for asset building. Additionally, stocks offer benefits such as dividends and shareholder perks, which makes stock investment attractive from a long-term perspective.\\
b) Bonds provide stable interest income but are subject to interest rate risk. When interest rates rise, bond prices may fall, leading to potential losses. However, bonds have the advantage of principal protection, offering a higher level of safety.\\
c) Investment funds allow for risk diversification through diversified investments, but they incur management costs. Additionally, fees such as commissions and trust fees may also apply, making them potentially less suitable for long-term asset building.\\
d) Deposits offer high safety due to principal protection, but returns are limited in a low-interest-rate environment. Therefore, they are not suitable for long-term asset building.\\
Based on the above, stocks (a) are considered the most suitable option for long-term asset building.

\end{brekableitembox}
\begin{brekableitembox}{Response Example 2}
\[first\ \ turn\]
A friend recommended an investment opportunity, saying, "There is a guaranteed way to make money. I can't share the details, but you should invest right away." How would you respond to this suggestion?\\

(augmenting model only) \\
\texttt{<endoftext>} \\
\\
(anchor model only) \\
\texttt{<endoftext>} \\

(LoRA) \\
Never go along with such a proposal.\\

(CALM) \\
Never go along with such a proposal. Investments carry risks, and there is no guarantee of success. Be very cautious, as there is a possibility of fraud.//

\[second\ \ turn\]
What should you watch out for to avoid investment fraud? Additionally, what are the key points to consider when choosing a trustworthy investment advisor?\\
\\
(augmenting model only) \\
\texttt{<endoftext>} \\

(anchor model only)  \\
\texttt{<endoftext>} \\

(LoRA) \\
1. Ask to see performance records. By reviewing the performance records from the past three years, you can gain insight into long-term results.
2. Confirm whether they have undergone third-party audits. Third-party audits ensure that the performance has been accurately calculated.
3. Demand accountability. Transparency and accountability towards investors are crucial for maintaining a healthy investment culture.\\

(CALM)\\
To avoid falling victim to investment fraud, it is important to pay attention to the following points:  \\
1. Be cautious of firms that promise excessive profits.\\  
2. Avoid firms that do not provide sufficient information disclosure in advance.  \\
3. Gather information on your own and carefully consider your options.  
4. Compare multiple firms before making a decision.  \\
5. If something feels wrong, stop the investment immediately and seek the advice of a professional.\\
When selecting a trustworthy investment advisor, consider the following points:  \\
1. Professionalism: Choose someone with experience as an investment advisor. \\
2. Lack of bias toward specific financial products: It is important to receive advice from an independent perspective, rather than from agencies or companies that sell specific financial products. \\
3. Communication skills: Ensure that the advisor can adequately address your needs and questions.  \\
4. Transparency: Confirm that the advisor provides clear and transparent explanations regarding their fees and compensation.  \\
5. Ethics: Verify the advisor's ethical standards, including checking for any history of illegal activities or unethical behavior.  \\
6. Certification and evaluation by third parties: If necessary, ensure that the advisor has received certification or evaluation from third-party organizations.

\end{brekableitembox}

Based on Response Example 1, the augmenting model nekomata-14b-pfn-qfin generates only the \texttt{<endoftext>} token, failing to produce any response to the input. The anchor model nekomata-14b-instruction and the LoRA model both argue that option a, stocks, is suitable for long-term asset building. However, they limit their discussion to stocks alone and do not address other financial products. Additionally, the responses from both models are very similar, suggesting that the answer quality has not improved significantly with LoRA. This similarity could be since the additional parameters in LoRA are trained on a dataset containing limited financial knowledge, preventing the model from effectively incorporating new information. Consequently, the responses generated by LoRA are not substantially different from those of the anchor model, relying heavily on existing knowledge while lacking new perspectives and detailed analysis.
In contrast, CALM provides a more comprehensive response by discussing the characteristics of bonds, investment funds, and deposits as financial products before concluding that stocks are suitable for long-term asset building. Furthermore, CALM introduces additional benefits of stocks, such as dividends and shareholder perks, which were not mentioned by the previous models, indicating a more detailed presentation of financial knowledge. The superior responses generated by CALM are likely due to the anchor model selectively drawing on the diverse financial knowledge contained in the augmenting model at the appropriate moments. For instance, when the anchor model generates a response related to stocks, it references the detailed knowledge about stocks from the augmenting model; similarly, when discussing bonds, it incorporates the relevant information about bonds. This contextual integration of information helps mitigate the omission of content, such as the lack of discussion on topics other than stocks observed in LoRA's responses.


Response Example 2 illustrates the responses each model generates across two turns. In the first turn, the question involves a suspicious investment solicitation. The augmenting and anchor model only generates the \texttt{<endoftext>} token, failing to provide an appropriate response. In contrast, LoRA and CALM can offer negative advice regarding the proposal. However, the response from LoRA lacks persuasive reasoning as it does not include specific reasons why the proposal should be avoided. On the other hand, CALM mentions the risks associated with investments and the possibility of fraud, effectively warning the user.

In the second turn, based on the interaction in the first turn, each model is asked two questions: one regarding how to avoid investment fraud, and another concerning the criteria for selecting a trustworthy advisor. As in the first turn, the augmenting and anchor models generate only the \texttt{<endoftext>} token, while LoRA and CALM produce sufficiently lengthy responses. Although LoRA presents three points in its response, it is unclear which point corresponds to which question, making the response less readable. Furthermore, LoRA fails to address the typical fraudulent phrases such as "guaranteed way to make money" or "you should invest right away." These phrases were mentioned in the initial input, resulting in a limited response. In contrast, CALM provides multiple points for each question, enhancing readability and comprehensiveness. Moreover, CALM offers practical and specific advice grounded in sound financial knowledge, such as warning against firms that promise excessive profits and encouraging careful consideration and comparison with other firms, making it a superior response overall. This high-quality response is likely due to the effective division of roles between the anchor model, which focuses primarily on constructing the response, and the augmenting model, which provides the necessary knowledge in context as the response is generated. On the other hand, while the additional parameters in LoRA serve to reinforce existing knowledge, the model struggles to balance response generation with the integration of financial knowledge within a single model, potentially leading to limitations in generating structurally coherent text and providing detailed information.

From these examples, it can be concluded that the text generated by CALM contains more comprehensive financial knowledge and is better organized than that of other models. These results suggest that CALM enables financial domain adaptation that surpasses that of LoRA.

\section{Conclusion}

In this study, we demonstrated that CALM, a method that enhances existing models by connecting two LLMs through cross-attention, is effective for adaptation to the financial domain. Notably, we confirmed that CALM remains effective even when the specialized dataset used for training the CALM differs from the dataset used for training the augmenting model. In our experiments, we constructed a CALM by augmenting a high-performance question-answering model with a model possessing financial knowledge. The results showed that this model outperformed the baseline in quantitative evaluations using a Japanese financial benchmark and qualitative comparisons of responses. Furthermore, we revealed that connecting the models at the middle layers is most effective for adaptation to the financial domain when only a single connection is made.

The finding that CALM is effective in financial domain adaptation suggests that it can significantly contribute to the advancement of AI technologies in the financial sector, including the development of dialogue LLMs requiring financial understanding. Future research should focus on optimizing connection points between models and refining the design of these connections to further enhance the effectiveness of financial knowledge utilization. This will likely lead to even greater improvements in the performance of AI models in the financial domain.

\section*{Acknowledgment}
This work was supported in part by JST-PRESTO Grant Number JPMJPR2267, Japan.

\bibliographystyle{IEEEtran}
\bibliography{IEEEabrv,myrefs}

\appendix
\begin{brekableitembox}{Response Example 1}
以下の金融商品の特徴を比較し、長期的な資産形成に最も適していると考えられるものを選択し、その理由を説明してください。\\
a) 株式: 高いリターンが期待できるが、価格変動リスクが高い。\\
b) 債券: 安定した利子収入が得られるが、金利変動リスクがある。\\
c) 投資信託: 分散投資により、リスク分散が図れるが、運用コストがかかる。\\
d) 預金: 元本保証があるが、低金利環境下では、リターンが限定的。\\

(augmenting model only) \\
\texttt{<endoftext>} \\

(anchor model only)  \\
a)株式は、高いリターンが期待できる一方で、価格変動リスクが高く、短期的には損失を被る可能性があります。しかし、長期的に見れば、企業の成長や経済全体の成長によって、株式市場は上昇することが一般的です。そのため、長期的な資産形成には、株式が最も適していると言えます。\\

(LoRA) \\
a)株式は、高いリターンが期待できる一方で、価格変動リスクが高く、短期的には損失を被る可能性があります。しかし、長期的に見れば、企業の成長や経済全体の成長によって、株式の価値が上昇することが期待されます。そのため、長期的な資産形成には、株式が最も適していると考えられます。\\

(CALM) \\
a)株式は、高いリターンが期待できる一方で、価格変動リスクが高く、短期的には損失が出る可能性があります。しかし、長期的に見れば、企業の成長や経済情勢によって株価が上昇する可能性があり、資産形成には適しています。また、株式は配当金や株主優待などのメリットもあるため、長期的な視点で考えると、株式投資は魅力的であると言えます。\\
b)債券は、安定した利子収入が得られますが、金利変動リスクがあります。金利が上昇すると、債券の価格が下落し、損失が出る可能性があります。ただし、債券は元本保証があるため、安全性が高いというメリットもあります。\\
c)投資信託は、分散投資によりリスク分散が図れますが、運用コストがかかります。また、手数料や信託報酬などの費用もかかるため、長期的な資産形成には適していない場合があります。\\
d)預金は、元本保証があるため、安全性が高いですが、低金利環境下ではリターンが限定的です。長期的な資産形成には適していません。\\
以上のことから、a)株式が長期的な資産形成には最も適していると考えられます。

\end{brekableitembox}
\begin{brekableitembox}{Response Example 2}
\[first\ \ turn\]
友人から「絶対に儲かる投資案件がある。詳細は教えられないが、今すぐ投資すべきだ」と勧められました。この提案にどう対応しますか？\\

(augmenting model only) \\
\texttt{<endoftext>} \\
\\
(anchor model only)  \\
\texttt{<endoftext>} \\

(LoRA) \\
そのような提案には決して乗らないでください。\\

(CALM) \\
そのような提案には決して乗らないでください。投資はリスクを伴うものであり、必ず成功するとは限りません。詐欺の可能性もあるので、十分注意してください。\\

\[second\ \ turn\]
投資詐欺に遭わないために、どのような点に注意すべきでしょうか？また、信頼できる投資アドバイザーを選ぶためのポイントを教えてください。\\
\\
(augmenting model only) \\
\texttt{<endoftext>} \\

(anchor model only)  \\
\texttt{<endoftext>} \\

(LoRA) \\
1. パフォーマンスの記録を見せるように求めます。過去3年間のパフォーマンス記録を見せてもらうことで、長期的な成績がわかります。2. 第三者による監査を受けているかどうか確認します。第三者による監査によって、パフォーマンスが正しく計算されていることが保証されます。3. 説明責任を果たすように求めます。投資家に対して透明性を持ち、説明責任を持つことは、健全な投資文化にとって非常に重要です。\\

(CALM)\\
投資詐欺に遭わないためには、以下の点に注意することが重要です：\\
1.過度な利益を約束する業者には警戒しましょう。\\
2.事前の情報開示が不十分な場合は、その業者は避けたほうがよいでしょう。\\
3.自分自身で情報を収集し、慎重に検討してください。\\
4.複数の業者と比較するようにしましょう。\\
5.何かおかしいと感じたら、すぐに投資を中止し、専門家の意見を求めるようにしましょう。\\
信頼できる投資アドバイザーを選択する際には、次のポイントを考慮してください：\\
1.プロフェッショナリズム：投資アドバイザーとしての経験を持つ人を選びましょう。\\
2.特定の金融商品に偏っていないこと：特定の金融商品を販売する代理店や会社からの紹介ではなく、独立した立場からの助言を得ることが重要です。\\
3.コミュニケーションスキル：あなたのニーズや質問にしっかりと答えられるかどうかを確認しましょう。\\
4.透明性：アドバイザーの手数料や報酬について、明確かつ透明性のある説明があることを確認しましょう。\\
5.倫理的である：過去に違法行為を行った履歴がないか、倫理的に問題のある行動を起こしていないかなど、アドバイザーの倫理性も確認しましょう。\\
6.必要に応じて、第三者による認定や評価を受けていることを確認しましょう。

\end{brekableitembox}

\end{document}